\documentclass[conference]{IEEEtran}
\IEEEoverridecommandlockouts
\usepackage{amsmath,amssymb,amsfonts, amsthm}
\usepackage{algorithmic, algorithm}

\usepackage{graphicx}
\graphicspath{ {./images/} }

\usepackage{textcomp}
\usepackage{xcolor}
\usepackage{mathrsfs}
\usepackage[numbers]{natbib}
\usepackage{comment}
\usepackage{caption}
\usepackage{subcaption}
\usepackage{hyperref}
\hypersetup{colorlinks}
\hypersetup{colorlinks,linkcolor={red},citecolor={olive},urlcolor={red}}

\usepackage{amsmath}

\DeclareMathOperator*{\argmin}{arg\,min}

\def\BibTeX{{\rm B\kern-.05em{\sc i\kern-.025em b}\kern-.08em
T\kern-.1667em\lower.7ex\hbox{E}\kern-.125emX}}
\begin{document}

\title{Applications of Online Nonnegative Matrix Factorization to Image and Time-Series Data}

\author{\IEEEauthorblockN{Hanbaek Lyu}
\IEEEauthorblockA{\textit{Department of Mathematics} \\
\textit{University of California}\\
Los Angeles, CA 90025 \\
hlyu@math.ucla.edu}
\and
\IEEEauthorblockN{Georg Menz}
\IEEEauthorblockA{\textit{Department of Mathematics} \\
\textit{University of California}\\
Los Angeles, CA 90025 \\
gmenz@math.ucla.edu}
\and
\IEEEauthorblockN{Deanna Needell}
\IEEEauthorblockA{\textit{Department of Mathematics} \\
\textit{University of California}\\
Los Angeles, CA 90025 \\
deanna@math.ucla.edu}
\and
\IEEEauthorblockN{Christopher Strohmeier}
\IEEEauthorblockA{\textit{Department of Mathematics} \\
\textit{University of California}\\
Los Angeles, CA 90025 \\
c.strohmeier@math.ucla.edu}
}

\date{February 2020}

\maketitle

\begin{abstract}
Online nonnegative matrix factorization (ONMF) is a matrix factorization technique in the online setting where data are acquired in a streaming fashion and the matrix factors are updated each time. This enables factor analysis to be performed concurrently with the arrival of new data samples. In this article, we demonstrate how one can use online nonnegative matrix factorization algorithms to learn joint dictionary atoms from an ensemble of correlated data sets. We propose a temporal dictionary learning scheme for time-series data sets, based on ONMF algorithms. We demonstrate our dictionary learning technique in the application contexts of historical temperature data, video frames, and color images.

\end{abstract}

\section{Introduction}


In the last few decades, the quantity of data available and the need to effectively exploit this data have grown exponentially. At the same time, modern data also presents new challenges in its analysis for which new techniques and ideas have become necessary. Many of these techniques may be classified as \textit{topic modeling} (or \textit{dictionary learning}), which aim to extract important features of a complex dataset so that one can represent the dataset in terms of a reduced number of extracted features, or topics. One of the advantages of topic modeling-based approaches is that the extracted topics are often directly interpretable, as opposed to the arbitrary abstractions of deep neural networks.

\begin{figure}[H]
\centering
\includegraphics[width=1 \linewidth]{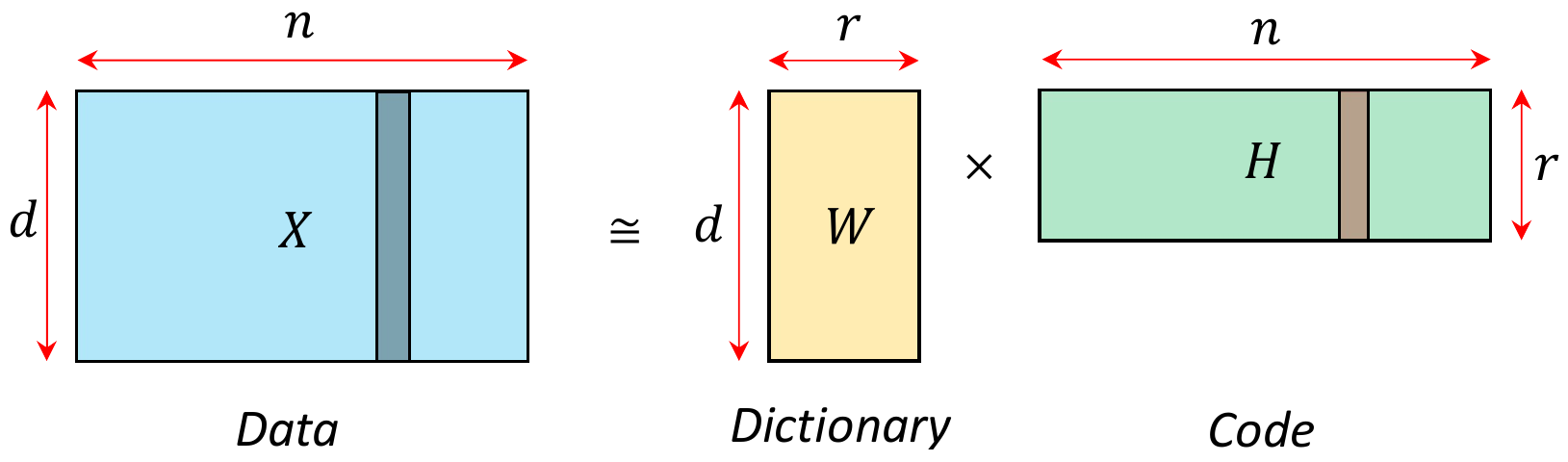}
\caption{Illustration of  matrix factorization. Each column of the data matrix is approximated by the linear combination of the columns of the dictionary matrix with coefficients given by the corresponding column of the code matrix.}
\label{fig:NMF_diagram}
\end{figure}

\textit{Matrix factorization} provides a powerful setting for dimensionality reduction and dictionary learning problems. In this setting, we have a data matrix $X \in \mathbb{R}^{d \times n}$, and we seek a factorization of $X$ into the product $WH$ for $W \in \mathbb{R}^{d \times r}$ and $H \in \mathbb{R}^{r \times n}$ (see Figure \ref{fig:NMF_diagram}). Hence each column of the data matrix is approximated by the linear combination of the columns of the \textit{dictionary matrix} $W$ with coefficients given by the corresponding column of the \textit{code matrix} $H$. This problem has been extensively studied under many names each with different constraints: dictionary learning, factor analysis, topic modeling, component analysis. It has also found applications in text analysis, image reconstruction, medical imaging, bioinformatics, and many other scientific fields more generally \cite{sitek2002correction, berry2005email, berry2007algorithms, chen2011phoenix, taslaman2012framework, boutchko2015clustering, ren2018non}.

In today's data world, large companies, scientific instruments, and healthcare systems are collecting massive amounts of data every day so that an entire data matrix is hardly available at any single time. \textit{Online matrix factorization} is a matrix factorization problem in the online setting where data are accessed in a streaming fashion and the matrix factors are updated each time. This enables factor analysis to be performed concurrently with the arrival of new data samples.

In this article, we demonstrate three application contexts of matrix factorization algorithms, namely, historical temperature data, color images, and video frames. In doing so, we demonstrate how one can use matrix factorization techniques to learn joint dictionary atoms from an ensemble of correlated data sets. In the case of historic temperature data, we show that the online-learned joint dictionaries reveal key features of dependence between temperatures of four cities (LA, SD, SF, and NYC), which we use to ``in-paint" missing data in one city by inferring from observed data in other cities. By a suitable matricization, we learn a small number of dictionary patches from a color image and reconstruct the original image from this low-rank image basis. The same procedure, used in conjunction with a convolutional neural network for classifying image patches, enables us to restore color to a grayscale image.  Lastly, for video frames, we compare dictionary frames learned from ``offline" and online matrix factorization algorithms and show that they capture different features of the video frames. By factorizing along the time dimension, we also demonstrate that one can detect a significant temporal change in the dataset using matrix factorization algorithms.

\section{Dictionary learning by online Nonnegative Matrix Factorization}

In this section, we describe a dictionary learning algorithm based on online nonnegative matrix factorization.

\subsection{Nonnegative Matrix Factorization}

\textit{Nonnegative matrix factorization} (NMF) is a variant of matrix factorization where one seeks find two smaller matrices whose product approximates a given nonnegative data matrix. Below we give an extension of NMF with extra sparsity constraint on the code matrix, which is often used for dictionary learning problems. Given a data matrix $X\in \mathbb{R}^{d\times n}_{\ge 0}$, the goal is to find nonnegative dictionary $W\in \mathbb{R}^{d\times r}$ and code matrices $H\in \mathbb{R}^{r\times n}$ by solving the following optimization problem:
\begin{align}\label{eq:NMF_error1}
\inf_{W\in \mathbb{R}^{d\times r}_{\ge 0},\, H\in \mathbb{R}^{r\times n}_{\ge 0} }  \lVert X - WH  \rVert_{F}^{2} + \lambda\lVert H \rVert_{1},
\end{align}
where $\lVert A \rVert_{F}^{2} = \sum_{i,j} A_{ij}^{2}$ denotes the matrix Frobenius norm and $\lambda \ge 0$ is the $L_{1}$-regularization parameter for the code matrix $H$.

A consequence of the nonnegativity constraints is that one must represent the data using the dictionary $W$ without exploiting cancellation. In practice, this forces the learned atoms to be ``atomic," or localized. A famous comparison was presented in \cite{lee1999learning}. There, the authors apply PCA and NMF to (vectorized) images of human faces. The (matricized) learned dictionary atoms from PCA resembled entire faces, while the matricized dictionary atoms from NMF resembled parts of faces, edges, etc. Furthermore, by enforcing a sparsity constraint on the code matrix, the learned dictionary elements have to appear sparsely in approximating each column of the data matrix. To make the dictionary components more localized and reduce the overlaps between them, one can also enforce sparseness on the dictionary matrix \cite{Hoyer2004} in the optimization problem for NMF as in \eqref{eq:NMF_error1}.

Many efficient iterative algorithms for NMF are based on block optimization schemes that have been proposed and studied, including the well-known multiplicative update method by Lee and Seung \cite{lee2001algorithms} (see \cite{gillis2014and} for a survey).


\begin{algorithmic}

\begin{algorithm}
\caption{NMF by Multiplicative Updates}

\STATE \textbf{Input:} $W_0, H_0, K$
\FOR{$k = 1,...,K$}
\STATE Update code:
\STATE $\qquad H_{ij}^{k + 1} \gets H_{ij}^k \frac{((W^k)^TX)_{ij}}{((W^k)^TW^kH^k)_{ij}}$
\STATE Update dictionary:
\STATE $\qquad W_{ij}^{k + 1} \gets W_{ij}^k
\frac{(X(H^{k + 1})^T)_{ij}}{(W^k H^{k + 1}((H^{k + 1})^T))_{ij}}$

\ENDFOR

\end{algorithm}

\end{algorithmic}

\subsection{Online Nonnegative Matrix Factorization}
\label{subsection:ONMF}

\textit{Online Matrix Factorization} (OMF) is an optimization problem where given a sequence $(X_{t})_{t\ge 0}$ of data matrices, one seeks to find sequence of dictionary and code matrices $(W_{t},H_{t})_{t\ge 0}$ which asymptotically minimizes a loss function of choice. If we have the additional nonnegativity constraint on the factors, the resulting problem is called the \textit{Online Nonnegative Matrix Factorization} (ONMF).

One of the most fundamental ideas in many algorithms for OMF is empirical loss minimization. Roughly speaking, the idea is to choose $W_{t-1}$ be to the minimizer of an empirical loss function up to time $t-1$, and when the new data matrix $X_{t}$ arrives, we find an improved dictionary $W_{t}$ in response to the empirical loss up to time $t$. However, as the empirical loss function is usually nonconvex and difficult to optimize, many successful algorithms for OMF exploit techniques such as block optimization, convex relaxation, and majorization-minimization. Below we present one of the most well-known OMF algorithms proposed in \cite{mairal2010online}. The superscript of matrices in the algorithm below denotes iterates.

\begin{algorithmic}

\begin{algorithm}
\caption{Online Nonnegative Matrix Factorization}

\STATE \textbf{Input:} $W_0, \lambda$
\FOR{$t = 1,...,T$}
\STATE Update sparse code:
\STATE \qquad$H_t = \argmin\limits_{H \geq 0} ||X_t - W_{t - 1}H||_F^2 + \lambda||H||_1$
\STATE Aggregate data:
\STATE \qquad$A_t = \frac{1}{t}((t - 1)A_{t - 1} + H_tH_t^T)$
\vspace{0.1cm}
\STATE \qquad$B_t = \frac{1}{t}((t - 1)B_{t - 1} + H_tX_t^T)$
\STATE Update dictionary:
\STATE \qquad $W_t = \argmin\limits_{W \geq 0}\frac{1}{2}tr(W A_t W^T) - tr(B_t W)$

\ENDFOR

\end{algorithm}
\end{algorithmic}

Rigorous convergence guarantees for online NMF algorithms have been obtained in \cite{mairal2010online} for independent and identically distributed input data. Recently, convergence guarantees of online NMF algorithms have been established when the data matrices have hidden Markov dependence \cite{lyu2019online}, ensuring further versatility of NMF based topic modeling from input sequences generated by Markov Chain Monte Carlo algorithms.

A simple computation reveals that the minimization problem in the update of the dictionary $W$ equivalant to the more intuitive problem,
\begin{align*}
W_t = \argmin\limits_{W \geq 0} \frac{1}{t}\sum_{s = 1}^{t} ||X_s - W H_s||_F^2 + \lambda ||H_s||_1
\end{align*}
where the objective function is a convex upper bounding surrogate of the corresponding empirical loss function. In practice, there is critical difference: the latter optimization problem requires one to have access to all of the data up to time $t$ in order to obtain $W_t$, while the former only needs the ``aggregation matrix" of the data and the data sample $X_t$. This property of the formulation of the algorithm is what enables ONTF to deal with the data size, acquisition, and evolution problems mentioned in the introduction.

We comment on a related important contrast between NMF and ONMF. While the NMF algorithm in the last subsection was essentially symmetric in the update of the dictionary and code for $\lambda = 0$, this is far from true in the ONMF algorithm. This is intuitive: in ONMF, one attempts to learn a fixed dictionary that can well-model all data in expectation. However one particular data sample generally has little to do with another, and so there is no reason for their respective codes to be similar.

\section{Time-series application}

In this section, we apply ONMF to time-series data for online dictionary learning and online reconstruction. We will be using the online nature of the dictionary learning algorithm there.

Suppose we observe a single numerical value $x_{s}\in \mathbb{R}$ at each discrete time $s$. By adding a suitable constant to all observed values, we may assume that $x_{s}\ge 0$ for all $s\ge 0$. Fix integer parameters $k,N,r\ge 0$. Suppose we only store $N$ past data at any given time, due to a memory constraint. So at time $t$, we hold the vector  $\mathcal{D}_{t}=[x_{t-N+1},x_{t-N+2},\cdots,x_{t}]$ in our memory. The goal is to learn dictionary patterns of $k$-step evolution from the observed history $(x_{s})_{0\le s \le t}$ up to time $t$. A possible approach is to form a $k$ by $N-k+1$ data matrix $X_{t}$, whose $i$th column consists of the $k$ consecutive values of $\mathcal{D}_{t}$ starting from its $i$th coordinate. We can then factorize this into $k$ by $r$ dictionary matrix $W$ and $r$ by $t-k$ code matrices using an NMF algorithm:
\begin{align}
X_{t}=
\begin{bmatrix}
x_{t-N+1} & x_{t-N+2} & \cdots &  x_{t-k+1} \\
x_{t-N+2} & x_{t-N+3} & \cdots & x_{t-k+2} \\
\vdots & \vdots & \vdots &  \vdots \\
x_{t-N+k} & x_{t-N+k+1} & \cdots & x_{t} \\
\end{bmatrix}
\approx
W H.
\end{align}
This approximate factorization tells us that we can represent any $k$-step evolution from our past data $(x_{s})_{t-N< s \le t}$ approximately by a nonnegative linear combination of the $r$ columns of $W$. Hence the columns of $W$ can be regarded as dictionary patterns for all $k$-step time evolution patterns in our current dataset $\mathcal{D}_{t}$ at each time $t$.

In order to extend the above `temporal dictionary learning' scheme into an online setting, we apply the ONMF algorithm given in Subsection \ref{subsection:ONMF}. This only requires us to store additional aggregate matrices $A_{t}\in \mathbb{R}^{r\times r}$ and $B_{t}\in \mathbb{R}^{r\times d}$. We denote the resulting dictionary matrix at time $t$ as $W_{t}$. Then the $r$ columns of $W_{t}$ are temporal dictionary for all $k$-step evolution  in the time series $(x_{s})_{0\le s\le t}$ up to time $t$.

\begin{figure}[h]
\centering
\includegraphics[width=0.5\textwidth]{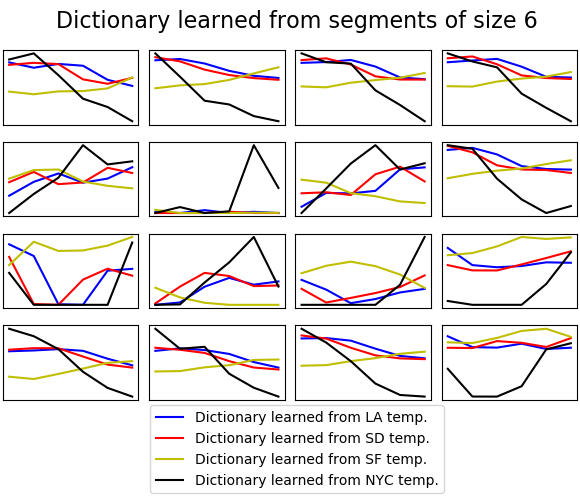}
\caption{16 joint temporal dictionary elements learned from the average monthly temperature of LA, SD, SF, and NYC during years 1944-2020. Each element indicates a fundamental pattern in the 6-month joint evolution of the temperatures in the four cities.
}
\label{fig:weather_dict}
\end{figure}

Lastly, we can further extend the above online temporal dictionary learning scheme for a collection of time-series data. We explain this in a concrete context. We analyze the historical monthly temperature (measured in Fahrenheit) data of four cities -- Los Angeles (LA), San Diego (SD), San Francisco (SF), and New York City (NYC), which is published as part of the \textit{NOAA online weather data (NOWData)} \cite{NOWData}. The time period of our data set spans the years 1944-2020, a total of 869 months. At each time step $0\le t\le 868$, we obtain four numerical values for the average monthly temperatures from the four cities, which we denote as a four-dimensional column vector $[LA(t), SD(t), SF(t), NYC(t)]^{T}$. Instead of applying our online temporal dictionary learning scheme separately for each city, we stack the resulting data matrices from the time series from each city vertically and form a joint data matrix at each time $t$. We use the parameters $k=6$, $N=50$, and $r=16$. Then we can learn a sequence $(W_{t})_{t\ge 0}$ of $4k$ by $r$ dictionary matrices using the ONMF algorithm. We plot the last dictionary $W_{868}$ after a suitable reshaping, which gives us 16 joint temporal dictionary elements as shown in Figure \ref{fig:weather_dict}.

The online-learned joint temporal dictionary elements shown in Figure \ref{fig:weather_dict} reveal an interesting dependency structure between the temperature evolution of the four cities. For instance, LA and SD (blue and red) evolve almost in synchrony with a small amplitude. On the other hand, NYC (black) has a larger amplitude and has a mostly positive correlation with LA and SD. Lastly, SF (yellow) has an intermediate amplitude with a negative correlation between LA and SD.

\begin{figure*}
\centering
\includegraphics[width=.9 \textwidth]{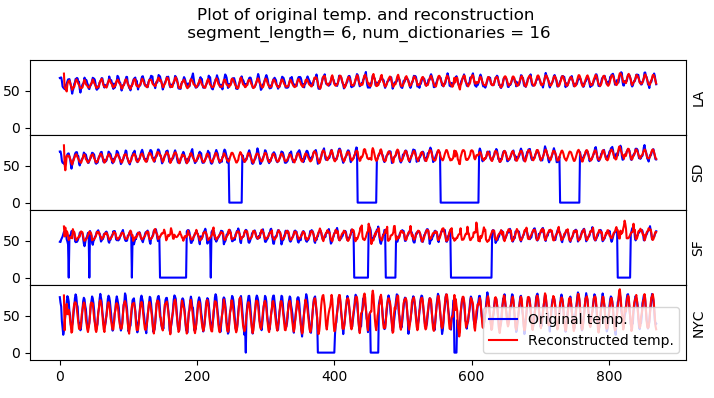}
\caption{(Blue) Plot of historical average monthly temperature (measured in Fahrenheit) data of four cities Los Angeles (LA), San Diego (SD), San Francisco (SF), and New York City (NYC). Missing entries are coded as 0 $^\circ$F. (Red) Online-reconstructed data using online-learned joint temporal dictionary.}
\label{fig:weather_reconstruction}
\end{figure*}

One way to evaluate the accuracy of learned dictionary patterns is to use them to reconstruct the original dataset, as shown in Figure \ref{fig:weather_reconstruction}. Here for each time $t$, we reconstruct the 6-step (6-month) joint time evolution vector $\mathbf{v}_{t}$ during the period $[t-5,t]$ using the online-learned joint dictionary $W_{t}$ from the observe data up to time $t$. This can be easily done by obtaining the best coefficient matrix $H$ such that $\lVert \mathbf{v}_{t}-W_{t}H \rVert + \lambda\lVert H \rVert_{1}$ is minimized, which is a convex quadratic problem. Then the product $W_{t}H$ gives the reconstruction for $\mathbf{v}_{t}$. We take the time-$t$-coordinates from this reconstruction to get the four red curves in Figure \ref{fig:weather_reconstruction}, plotted on top of the original temperature data in blue.

We emphasize that there is no a priori training. Initial dictionary matrix $W_{0}$ is completely random, but we progressively learn better dictionary matrix $W_{t}$ over time. Also, we remark that for the three cities SD, SF, and NYC, there are several missing entries, which were coded as -100 $^\circ$F. For a plotting purpose, we have modified these missing entries to 0 $^\circ$F, which are shown as horizontal bars in the blue curve at height 0 in Figure \ref{fig:weather_reconstruction}. Our reconstruction (red) matches the original data (blue) accurately for the observed entires. Furthermore, we used a modified reconstruction algorithm so that we only approximate observed entries, and use the resulting linear coefficients to ``fill-in" the missing entries using our joint dictionary. The result indicates that we can use our online joint dictionary learning to inferring missing entries in our dataset, purely based on learning the dependency structure.

\section{Video Application}

In the previous section, we applied ONMF to two important problems in image processing which did not use any particular advantage of the algorithm, which serves to demonstrate its utility. In this section, we turn to applications of ONMF to video processing. It is here that the sequential nature of the ONMF algorithm enforces a qualitatively different structure on the learned dictionary atoms when compared to standard offline NMF.

We now describe our experiment. Our data consisted of a grayscale video of a candle, which consisted of 75 time frames. Each time frame was represented by an 80 x 30 matrix of pixel intensities.

\begin{figure}
\centering
\includegraphics[width=0.41\textwidth]{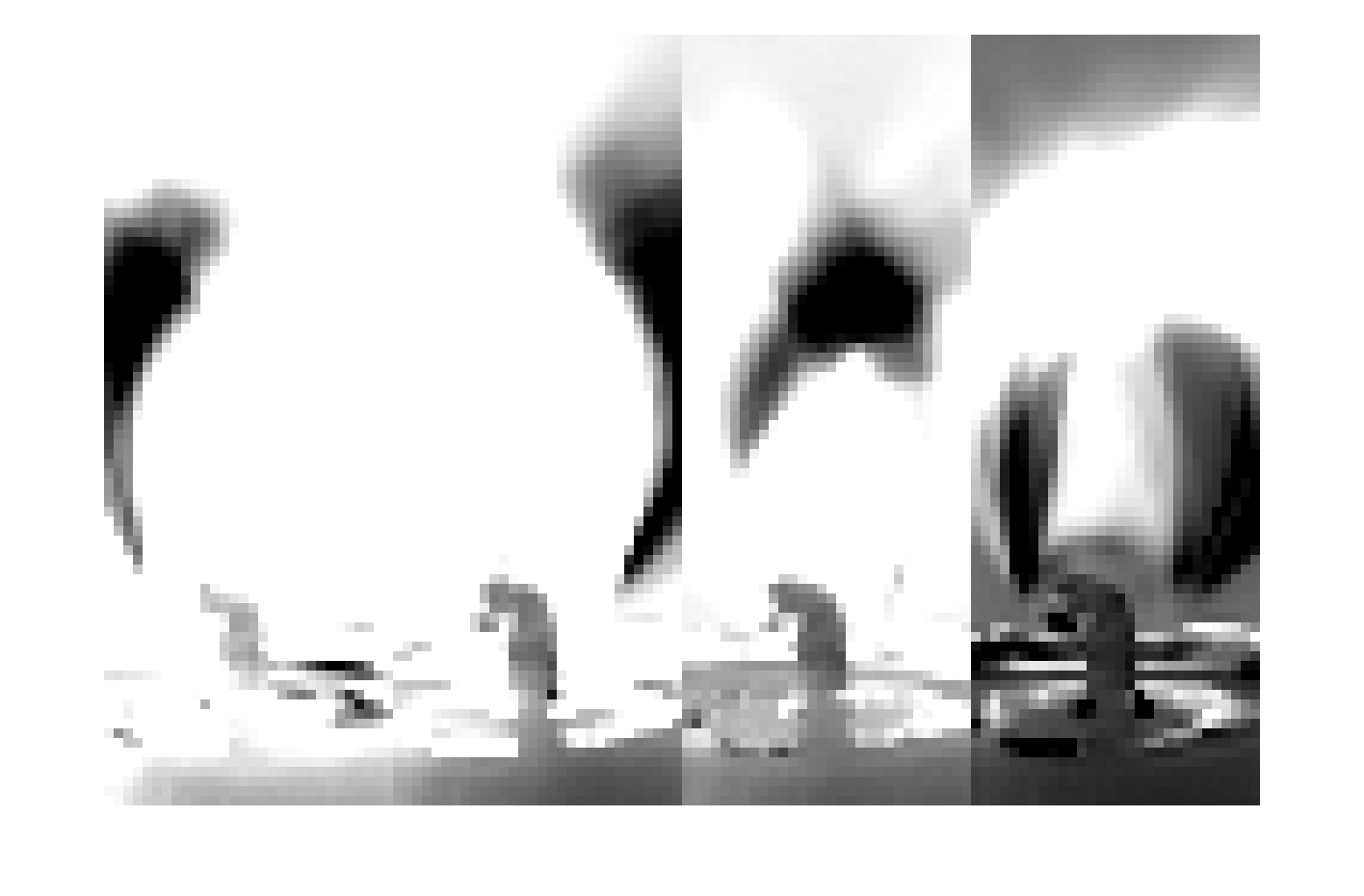}
\includegraphics[width=0.35\textwidth]{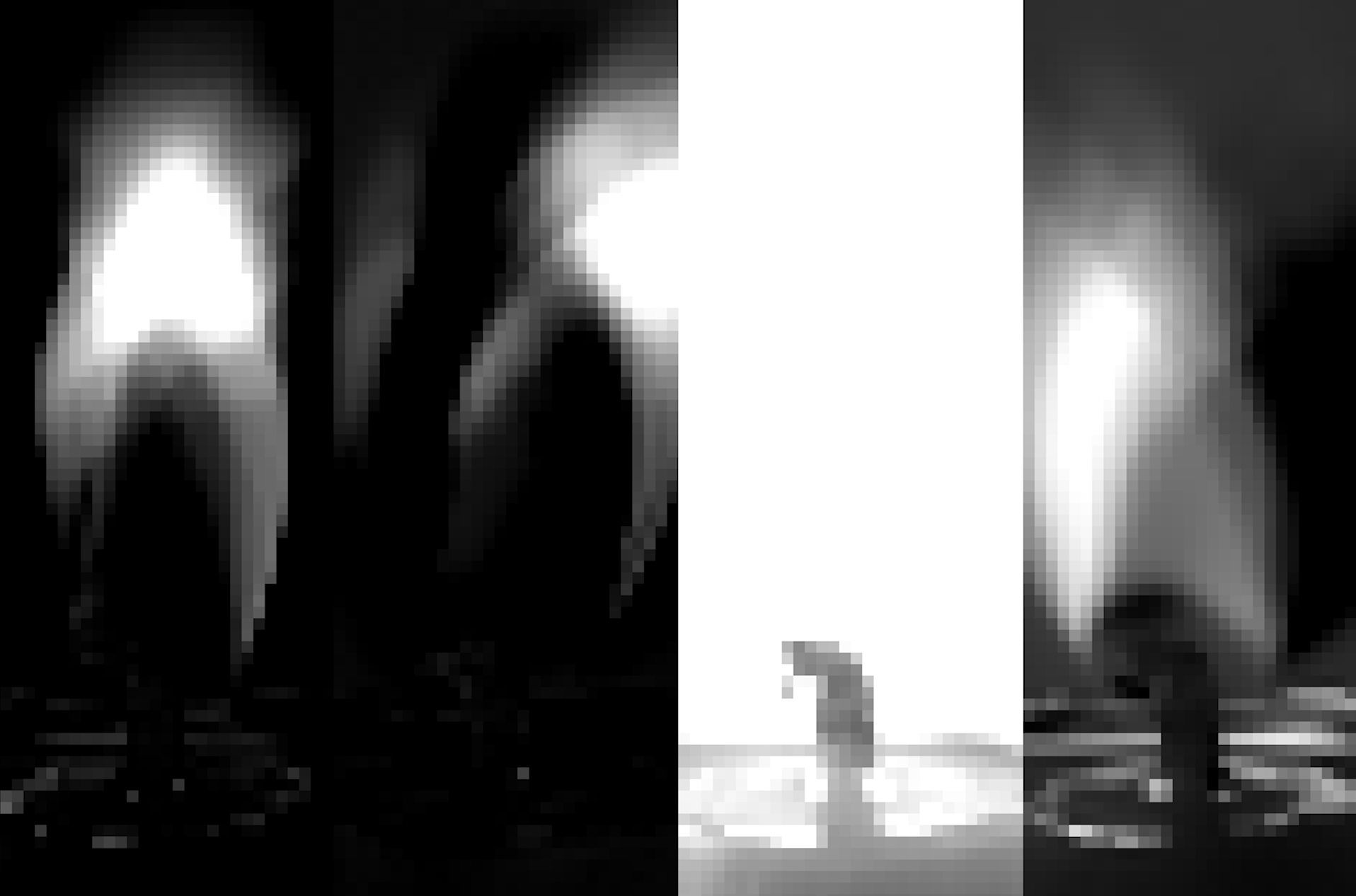}

\caption{Candle Video Dictionaries. The first dictionary consists of four elements and was trained by an alternating least squares-based, offline NMF, the second dictionary below was trained using ONMF, where each time frame of the video represented a new data point.}
\label{fig:candle dictionaries}
\end{figure}

We reshaped our data by vectorizing each time frame and then concatenating these vectors to form a single 2400 x 75 data matrix of pixel intensities. We applied an alternating least squares-based offline NMF to learn four dictionary elements. Next, we applied ONMF in natural way: each vectorized time frame corresponded to a new data sample. Again we learned four dictionary elements, which were then matricized and displayed in \ref{fig:candle dictionaries}.

\begin{figure}
\centering
\includegraphics[width=0.4\textwidth]{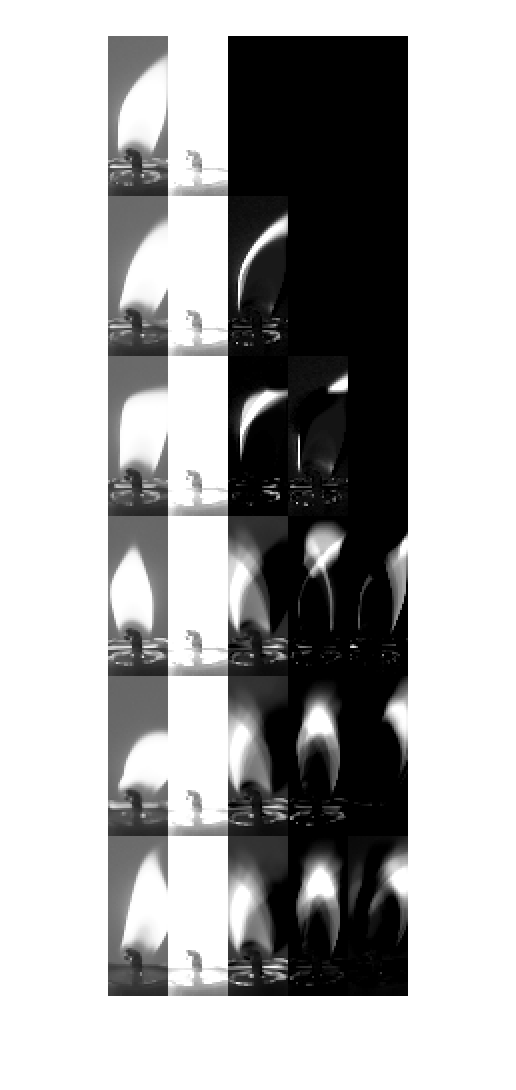}

\caption{Candle video and learned dictionary at various time frames  (time goes from top to bottom).  The left column corresponds to the actual video frame. The remaining four columns each correspond to a particular dictionary element. The six rows correspond to different time frames, 1, 5, 7, 15, 35, and 75}
\label{fig:video}
\end{figure}

Additionally, we plotted the dictionaries learned by ONMF up to various time frames and displayed the results. This reveals an interesting, and intuitive, illustration of how ONMF uncovers temporal structure in the video. In particular, like the candle in the video moves, initially trivial atoms one by one become nontrivial, and adopt the shape of a particular configuration of the candle, e.g. left, center, right. When one plots the full evolution of the learned dictionaries, once the basic patterns have been established we empirically observe that when the candle is in a certain configuration at a particular time frame, only the atom with the corresponding configuration exhibits any significant update. By the end of the learning process, the final atoms look like superpositions of small perturbations of a single configuration (e.g. the second atom resembles a superposition of right configurations, the fourth atom resembles a superposition of left configurations). The results are displayed in \ref{fig:video}.

\begin{figure*}
\centering
\includegraphics[width=1 \textwidth]{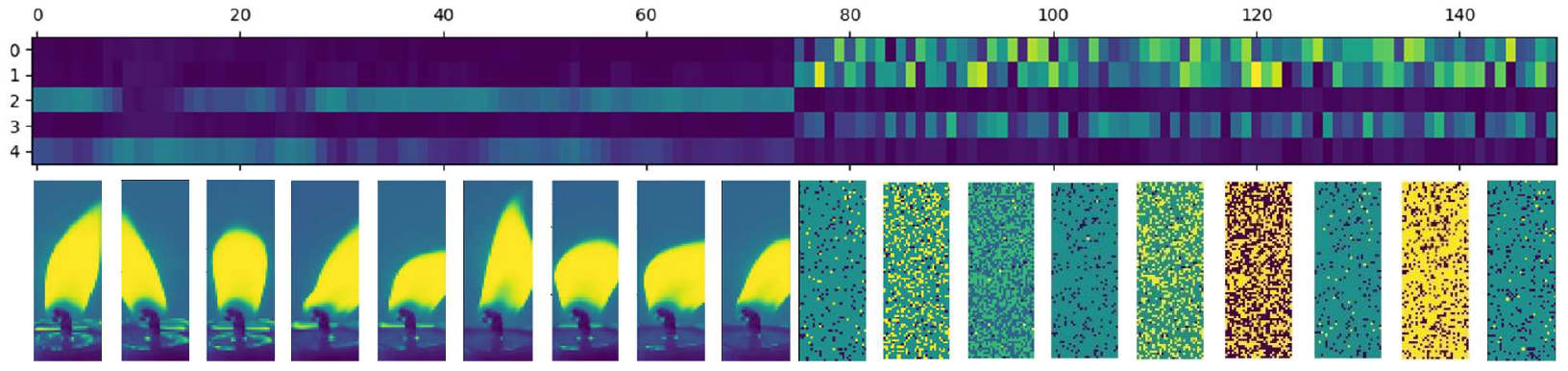}
\caption{Learning time evolution dictionary from a video frame using NMF. The first and last 75 frames of the video are from a candle video and white noise, respectively, as shown below. By an approximate factorization of shape [time x space] = [time x 5] [5 x space], we learn 150 x 5 dictionary matrix, whose columns give an approximate basis for the time evolution of each pixel of the video frame. The learned time evolution dictionaries detect the planted `phase transition' between frames 75 and 76.}
\label{fig:cancle_dict_time}
\end{figure*}

Lastly, we apply NMF to the video frame along the temporal dimension to learn the dictionary for time evolution. This will enable us to detect a temporally significant change in the video frame through the learned dictionary elements. To explain our simulation setup, first recall that the candle video frame is stored as an 80 x 30 x 75 tensor, where the last dimension corresponds to time. By flattening the first two spatial dimensions, this video frame can be represented as a 2400 x 75 matrix. In the previous experiments, we have factorized this data matrix using NMF and ONMF algorithms. Here, we take the transpose of this data matrix so that the dimension of 75 x 2400 correspond to [time x space]. We generate another video frame with the same size and length using white noise and concatenate with the candle video. Altogether, we form a data matrix $X$ of shape 150 x 2400, where the first 75 rows are coming from the candle video, and the last 75 are from white noise.

Next, we approximately factorize the data matrix $X$ into $WH$ using an NMF algorithm, where $W$ is of shape 150 x 5. This factorization tells us that each column of $X$, which corresponds to the time evolution of a single pixel for 150 frames, can be approximated by a nonnegative linear combination of the 5 columns of $W$. Hence the columns of $W$ give a dictionary for the time evolution of each of the 2400 pixels in the video frame. Since we have planted a significant change between times 75 and 76, these five time-evolution dictionary elements should be able to detect this ``phase transition" in our video. Indeed, as in Figure \ref{fig:cancle_dict_time}, all five columns of $W$ (shown as rows) exhibit significant change between frames 75 and 76, as expected. Furthermore, the first halves of the third and last dictionary elements in Figure \ref{fig:cancle_dict_time} contains information on the time evolution for the candle video, which disappears after the phase transition at frame 76. This is also expected since a basis for time evolution in the candle video should not contain meaningful information for the frames generated by white noise.

\section{ONMF for Color Image Processing}

In this section, we describe how ONMF can be applied to the two important image processing problems.

First, a quick introduction to patch-based image processing. A typical e.g. jpeg grayscale image may be represented by a matrix of unsigned integers with values between 0 and 255, representing the intensity. Lower values correspond to black, higher values to white. Color images may be stored similarly: to each of the three color channels red, blue and green, there corresponds an analogous matrix of color intensities.

In general, images may contain thousands of pixels, so working with the full image directly can be computationally infeasible, even when using online algorithms. Thankfully, one can achieve success on various image processing applications, such as compression, denoising, and impainting, by taking a ``patch-based" approach. The idea is to extract (typically, overlapping) small patches, usually of size in the range $8 \times 8$ to $30 \times 30$ pixels. One may apply some procedures to each patch and recover the full image by patch averaging.

\subsection{Color Image Compression}

In this subsection, we describe how one can utilize ONMF to implement a patch-based approach to image compression. These applications do not leverage any particular advantage of ONMF compared to other nonnegative matrix factorization algorithms, but serve to illustrate its versatility. However, in the video processing application we will have seen that ONMF can produce qualitatively distinct dictionaries compared to those obtained from standard offline nonnegative matrix factorization algorithms.

First, we consider the patch-based compression of images. The motivation for the approach presented here is as follows. A typical image, in its uncompressed form, is very high dimensional. Indeed, images of moderate size may contain thousands to millions of pixels. Thankfully, most real-world images are sufficiently regular to enable a substantial reduction in the effective dimension. For instance, while it is true that an image of a clear sky may contain a large number of pixels, there is little variation in this image, and so intuitively one may expect a representation by an object of dimension far lower than the number of pixels to be possible (indeed, a single blue image patch ought to suffice).

One of the most popular ways of exploiting the regularity of natural images is through the use of small image patches. The general scheme is to first learn a collection of relatively few ``atomic image patches" through some dictionary learning algorithms such as KSVD or NMF. One then may then implement one's application of choice, e.g. compression, denoising, inpainting, to some collection of overlapping small image patches that cover the full image. Finally, one can recover a full-size image through patch averaging.

\begin{figure}
\centering
\includegraphics[width=0.45\textwidth]{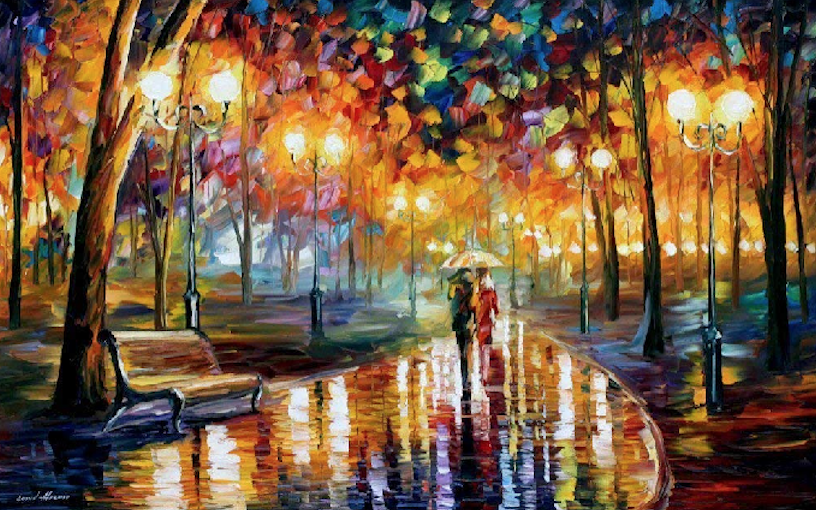}
\includegraphics[width=0.5\textwidth]{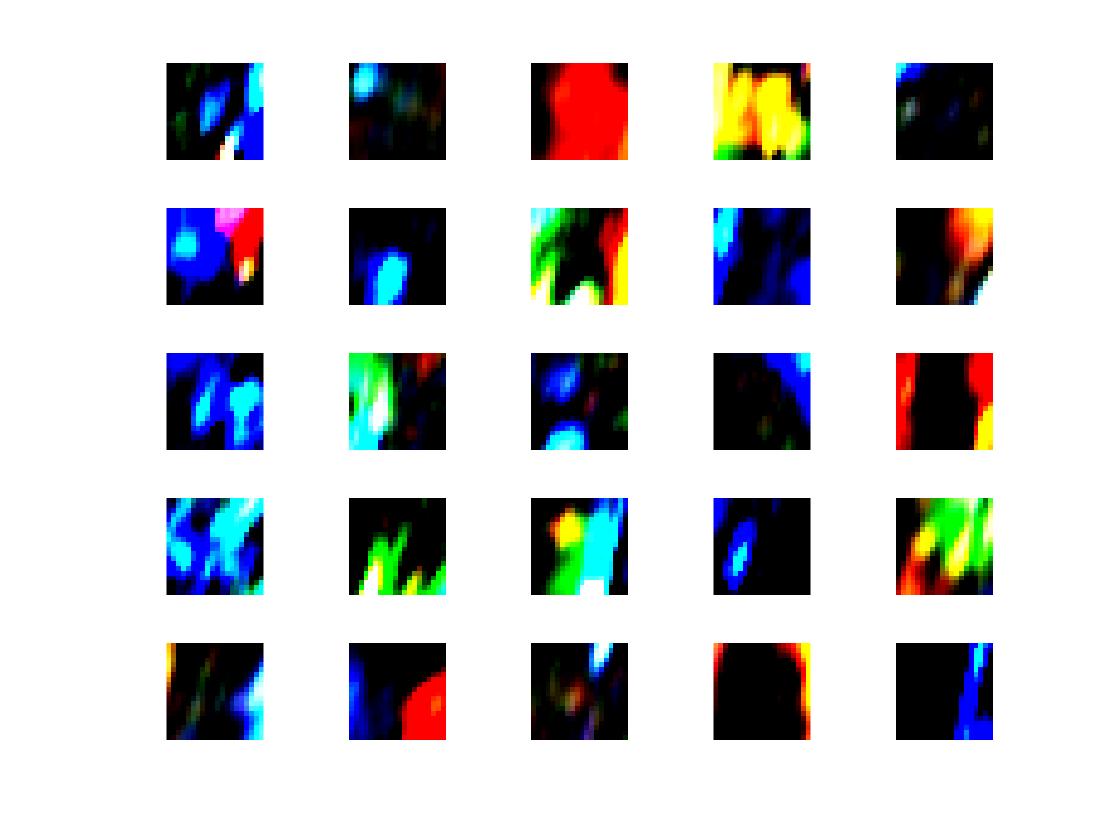}
\includegraphics[width=0.45\textwidth]{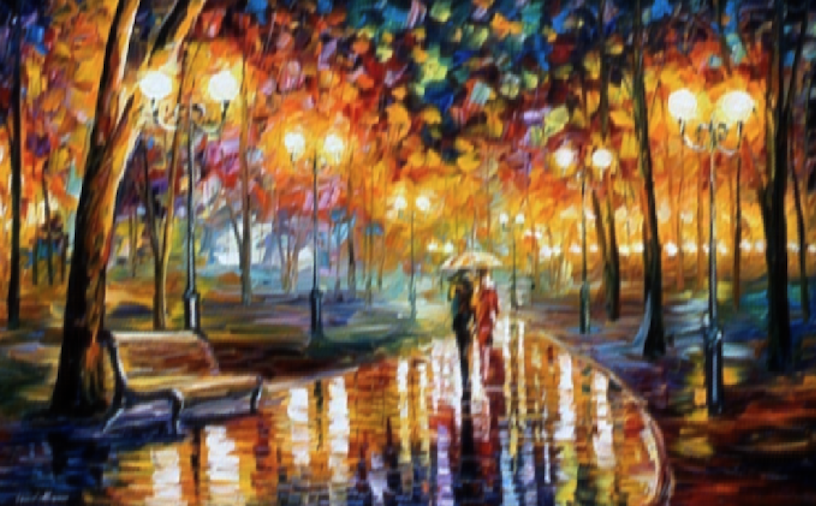}

\caption{Image Compression Via ONMF. (Top) uncompressed image of Leonid Afremov's famous painting ``Rain's Rustle." (Middle) 25 of the 100 learned dictionary elements, reshaped from their vectorized form to color image patch form. (Bottom): Painting compressed using a dictionary of 100 vectorized $20 \times 20$ color image patches obtained from 30 data samples of ONMF, each consisting of 1000 randomly selected sample patches. We used an overlap length of 15 in the patch averaging for the construction of the compressed image.}
\label{fig:painting}
\end{figure}

We now describe in more detail how we used ONMF to compress color images. ONMF sequentially receives data and updates the learned dictionary to effectively model new data samples. In our implementation, a single data sample will consist of a matrix whose columns consist of vectorized color image patches. For instance, we may select 1000 20 x 20 color image patches so that our data sample is a 1200 x 1000 matrix. The first 400 entries of the jth column are obtained by extracting the red channel of the jth image patch and vectorizing along the first axis. The next 400 entries are analogously obtained from the blue channel and the final 400 from the green channel. We may then apply ONMF as described in section II with some number of data samples. Figure \ref{fig:painting} displays the results of ONMF applied to a famous painting.

We comment that for quality compression, it is often necessary to use the significant overlap of patches. Too small overlap can lead to a certain ``blockiness" of the recovered image. For more information, consult the references \cite{Juvonen2017PatchbasedIR}, \cite{alkinani2017patch}, \cite{elad2010sparse}.

\subsection{Color Restoration}

Next, we describe how ONMF can be used in conjunction with convolutional neural networks to approach the problem of restoring color to grayscale images.

Color restoration without additional qualification of the term         ``restoration" is an ill-posed problem. Indeed, any sensible conversion from color to grayscale images is necessarily lossy. That said, of primary interest, are real-world images. For the sake of concreteness, one may have an old black and white photograph of a landscape, and wish to restore color to this photo. It is reasonable to assume a great degree of similarity between elements of old and recent photographs. Naturally occurring objects like grass, water, mountains, etc. look similar regardless of period. Thus the following idea suggests itself: learn a ``landscape dictionary," trained on color images of modern landscapes, and use this dictionary to restore color to the grayscale landscape image.

\begin{figure}
\centering
\includegraphics[width=0.25\textwidth]{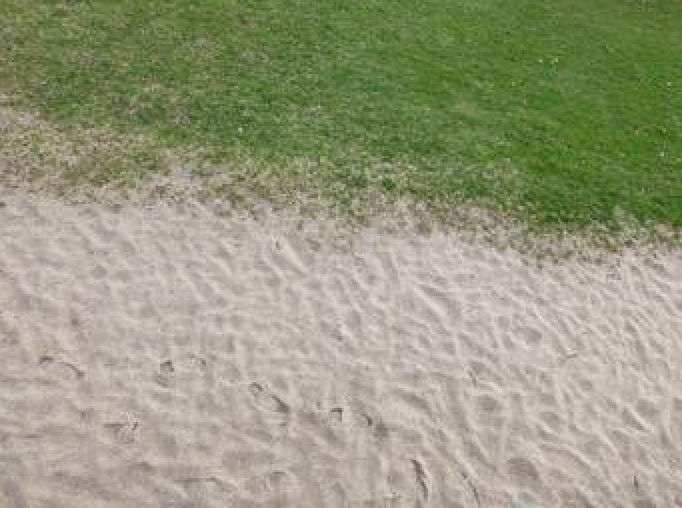}
\includegraphics[width=0.25\textwidth]{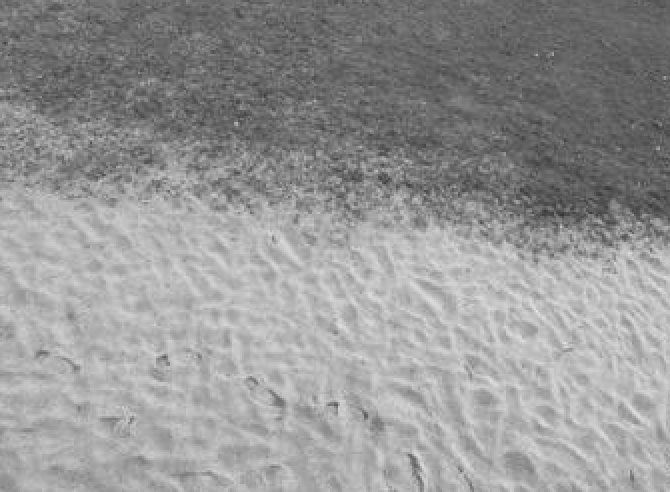}
\includegraphics[width=0.25\textwidth]{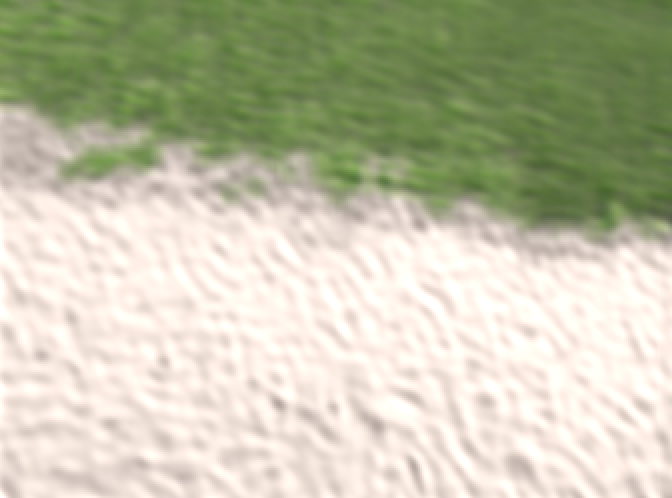}
\includegraphics[width=0.4\textwidth]{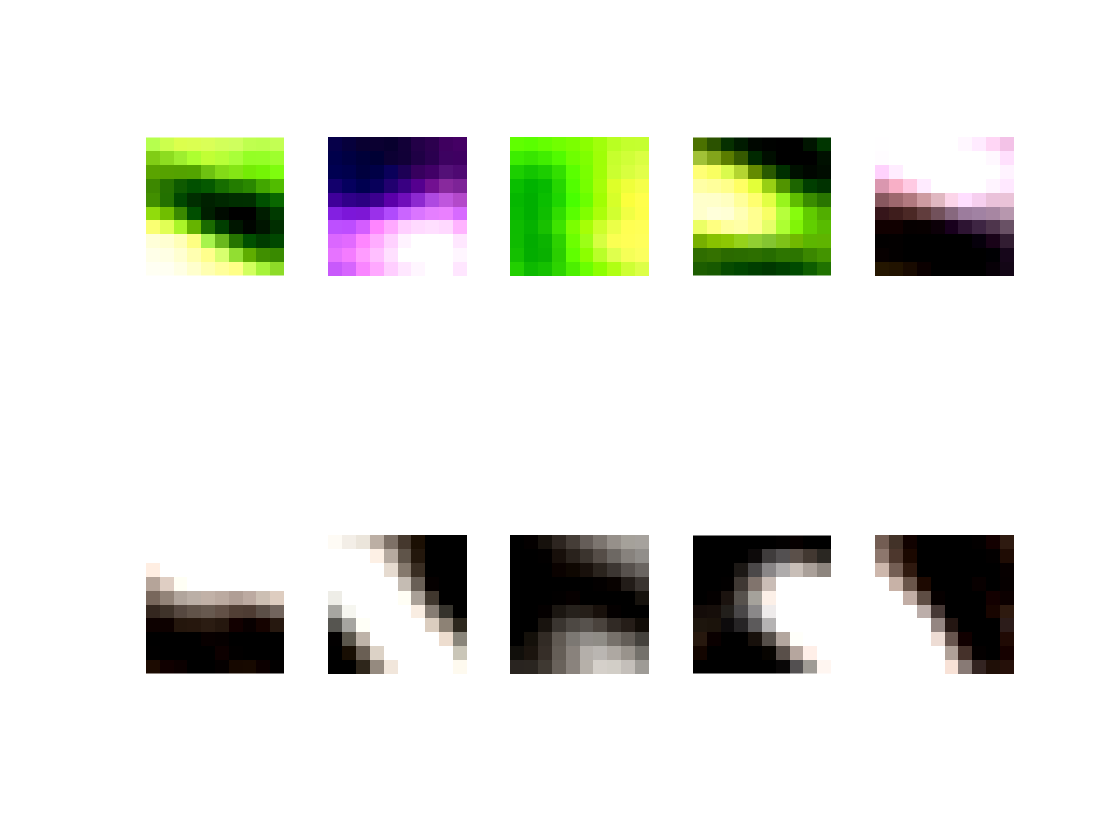}

\caption{Color Restoration. (First) Original image of grass and sand. (Second) Conversion of first image using Matlab's default (linear) color to grayscale conversion function, rgb2gray. (Third) Restored color image, obtained by applying our method to the second (grayscale) image using the dictionary below. (Fourth) The first row displays the grass dictionary, the second row displays the sand dictionary. Each dictionary contains five 10 x 10 color image patches. These were trained on our ONMF with 20 batches of 1000 randomly selected sample patches each. We used maximal patch overlap in the restoration process.}
\label{fig:grass and sand}
\end{figure}

One way to attempt the color restoration process is to take the landscape dictionary of color image patches and convert said patches to grayscale. There are many ways of doing so, but it is important to choose a linear conversion. With this dictionary of grayscale patches, one can use nonnegative least squares to approximate the image patches of the black and white landscape photo we wish to restore. The key idea is to use the coefficients in the representation by grayscale patches to form a color image patch using the corresponding color image dictionary elements. The linearity of the conversion from color to grayscale which produced the grayscale dictionary patches ensures that this procedure is consistent.

For this color restoration algorithm to be successful, it is necessary that in the above representation of an arbitrary grayscale image patch of the black and white photo, only (grayscale conversions of) the color landscape dictionary elements associated with this patch make a significant contribution. For example, if a grayscale image patch to be restored is extracted from a grass portion of the image, one would hope that ocean, mountain, cloud, etc. patches do not contribute. This may seem plausible at first, since the textures of all of these objects vary significantly, and so one might hope that in general only grayscale conversions of color grass image patches are used to represent black and white grass image patches. Unfortunately, this is not the case.

There are many difficulties in using a pure dictionary learning-based approach to color restoration. On one hand, images may abruptly change from pixel to pixel between different components of an image, e.g. the transition from mountain to sky. This suggests that one needs to use relatively small image patches, perhaps of size 10 x 10 or smaller, to avoid ``mountain patches" spilling into the sky portion of the image or vise-versa. On the other hand, if there is any hope for the necessary condition for the success of color restoration that different components of the black and white image may be identified despite their absence of color, one needs e.g. spatial features such as texture to distinguish grass from other components. For small very image patches, these features are difficult to detect. One can see from the dictionary atoms in both of our color image patch dictionaries that there is little textural information. Even larger image patches exhibit this problem: the low-rank approximation of arbitrary image patches enforces a degree of ``generality" in the few dictionary atoms learned to represent said patches and thus finer details like texture are obscured.

On the other hand, dictionary learning has its advantages. It takes relatively little data to learn an effective dictionary for tasks such as compression, and even color restoration of monochromatic objects. Motivated by this, we approached the color restoration problem in two stages. The first stage is to classify grayscale image patches of the black and white image to which we wish to restore color. Once the patch has been classified, we can use an appropriate dictionary of color image patches to restore color. For example, to restore color to black and white photos of landscapes, one may train a convolutional neural network to classify grayscale image patches as derived from grass, cloud, sky, mountain, water, etc. One can then train separate color image patch dictionaries corresponding to each of these elements.

We implemented this approach in a simple example of an image of grass juxtaposed with sand. The results of our algorithm are displayed in \ref{fig:grass and sand}. We comment that extensions of this idea like those described above are possible, although they require more effort.

\section{Conclusion}

In this paper, we have presented various applications of ONMF to image and video processing as well as to the analysis of time-series. Our experiments demonstrate both the utility of ONMF on applications of general interest as well as the special insights its online nature can reveal in sequential data.

\section*{acknowledgments}

Needell and Strohmeier were partially supported by NSF CAREER \#1348721 and NSF BIGDATA \#1740325.

\bibliographystyle{amsalpha}
\bibliography{ITA_bibliography.bib}

\end{document}